\documentclass{egpubl}
\usepackage{eg2021}

\ConferencePaper        
\usepackage[T1]{fontenc}
\usepackage{dfadobe}  

\usepackage{cite}  
\BibtexOrBiblatex
\electronicVersion
\PrintedOrElectronic
\ifpdf \usepackage[pdftex]{graphicx} \pdfcompresslevel=9
\else \usepackage[dvips]{graphicx} \fi

\usepackage{egweblnk} 

\usepackage{subcaption}
\title[Learning Human Search Behavior from Egocentric Visual Inputs]{Learning Human Search Behavior from Egocentric Visual Inputs}

\author[M. Sorokin \& W. Yu \& S. Ha \& C. K. Liu]
{\parbox{\textwidth}{
\centering
\vspace{-0.9cm}
Maks Sorokin$^{1}$\orcid{0000-0001-5994-0046}
\quad Wenhao Yu$^{1,2}$\orcid{0000-0001-8263-8224}
\quad Sehoon Ha$^{1,2}$\orcid{0000-0002-1972-328X}
\quad C. Karen Liu$^{3}$\orcid{0000-0001-5926-0905}
}
\\
{\parbox{\textwidth}{
\centering{
\vspace{-0.4cm}
\{maks,wenhaoyu,sehoonha\}@gatech.edu, karenliu@cs.stanford.edu \\
$^1$ Georgia Institute of Technology, Atlanta, GA, 30308, USA \\
$^2$ Robotics at Google, Mountain View, CA, 94043, USA \\
$^3$ Stanford University, Stanford, CA, 94305, USA
}
}
}
\vspace{-0.9cm}}

\usepackage{algorithm}
\usepackage[noend]{algpseudocode}
\usepackage{amsfonts}
\usepackage{hyperref}

\usepackage{xcolor}
\newcommand{\cmt}[1]{}

\long\def\ignorethis#1{}

\newcommand{\etal}{{\em{et~al.}\ }}

\newcommand{\ie}{i.e.\ }

\newcommand{\vc}[1]{\ensuremath{\mathbf{#1}}}

\newcommand{\mat}[1]{\ensuremath{\mathbf{#1}}}

\newcommand{\pctab}{\hspace{0.2in}}

\begin{document}

\teaser{
 \includegraphics[width=\linewidth]{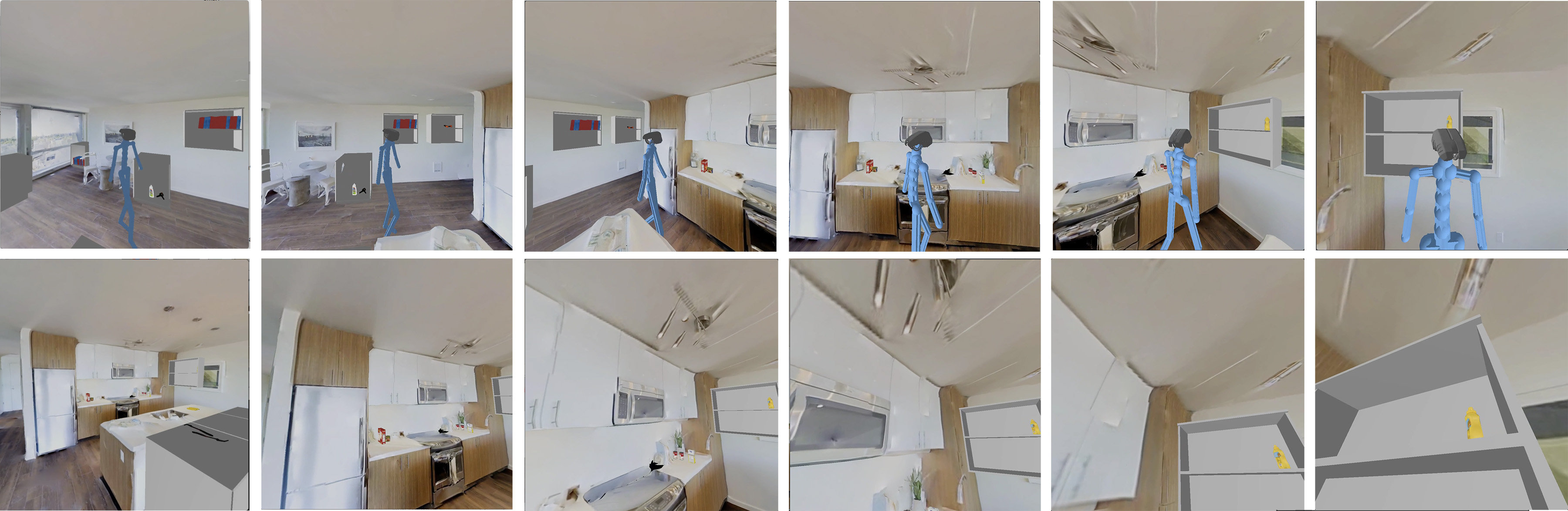}
 \centering
  \caption{A humanoid character learns to navigate and search for a target object (the mustard bottle) in a photorealistic 3D scene using its own egocentric vision and locomotion capability. Top: third-person view. Bottom: first-person view.}
\label{fig:teaser}
}

\maketitle
\begin{abstract}
``Looking for things'' is a mundane but critical task we repeatedly carry on in our daily life. We introduce a method to develop a human character capable of searching for a randomly located target object in a detailed 3D scene using its locomotion capability and egocentric vision perception represented as RGBD images. By depriving the privileged 3D information from the human character, it is forced to move and look around simultaneously to account for the restricted sensing capability, resulting in natural navigation and search behaviors. Our method consists of two components: 1) a search control policy based on an abstract character model, and 2) an online replanning control module for synthesizing detailed kinematic motion based on the trajectories planned by the search policy. We demonstrate that the combined techniques enable the character to effectively find often occluded household items in indoor environments. The same search policy can be applied to different full body characters without the need of retraining. We evaluate our method quantitatively by testing it on randomly generated scenarios. Our work is a first step toward creating intelligent virtual agents with humanlike behaviors driven by onboard sensors, paving the road toward future robotic applications.
\\


\end{abstract}

\section{Introduction}


``We spend about 5000 hours of our lives looking for things around the home'' \cite{ikea}. Indeed, searching for objects in complex indoor environments is a frequent event in our daily life--we look for ingredients in the kitchen for a recipe, we locate grocery items on the shelves in a supermarket, and we seem to always be in search of phones, keys or glasses many times a day. The goal of this paper is to model such important and ubiquitous behaviors by developing a virtual human capable of using its egocentric vision perception and locomotion capability to search for any randomly placed target object in a complex 3D scene. 

Search behaviors depend on simultaneous locomotion and survey of the environment, requiring modeling not only the physical motor skills, but also human sensory and decision making. Conventional character animation assumes full knowledge of the 3D environment and utilizes it to achieve optimal movements. While optimality is indeed observed in many human locomotion and manipulation tasks, it is also at odds with the stochastic nature of human sensing and decision making capabilities. Having the “oracle ability” to know exactly the 3D position of every vertex in the scene will likely to result in an ``optimal but unnatural'' search behaviors. Alternatively, imitating or tracking motion capture trajectories directly can potentially lead to natural human behaviors. However, pre-scripting or pre-planning a reference trajectory can be challenging for a search controller tasked to find objects placed at random locations in random scenes.

This paper builds on the hypothesis that equipping the virtual character with human-like sensing capabilities can lead to more natural behaviors, as demonstrated by previous work \cite{Yeo:2012,nakada2018deep,Eom:2020}. In particular, we limit the virtual character to egocentric vision perception from RGBD images when performing the search task. Without the full state information about the environment and the global position and orientation of itself, the character is forced to coordinate its motor capabilities to navigate and scan the scene to find the target object, naturally inducing humanlike decision-making behaviors under partial observation. Our method consists of two components: 1) a search control policy that determines where the character should move to and look at based on an abstract model, and 2) an online replanning control module for synthesizing the detailed kinematic motion based on the planned trajectories from the search policy. This decoupling of the task provides several benefits: first, training the search policy with an abstract model that only includes a torso and a head/camera is more computationally efficient for high-dimensional and large observation space, and second, a trained search policy can be re-used for multiple characters that share the same abstract model without retraining.

We use deep reinforcement learning (DRL) to train the search policy that takes as input the visual perception from the abstract model and predicts where it should move to and look at. Our training framework shares some similarities with existing works in visual navigation while having a few key distinctions. Unlike most vision-guided navigation applications, our agent can actively choose where to look independent of the body moving direction, enabling more plausible head movements and effective search strategies for the character. However, decoupling the body movement with the gaze direction makes the learning problem more challenging due to the larger observation space and the higher dimensional action space. We show that by combining Soft Actor Critic (SAC) with a constrastive proposed by Srinivas \etal~\cite{srinivas2020curl}, a vision-guided policy can effective learn the features from high dimensional pixels for our search problem. In addition, transferring the policy trained for an abstract model to a full-body character presents many challenges. Drawing analogy from sim-to-real transfer learning, we propose a  zero-shot online replanning method to transfer a model-agnostic policy to the biped human model and a wheel-based robot model. Combining offline visuomotor policy learning with online trajectory planning results in a virtual human capable of making motion plans using egocentric vision perception. 

We demonstrate our method on a human and a robotic character searching household items in realistic indoor scenes. We show that the character is able to find a small object, such as a pair of glasses in a large space including an open kitchen and a living room, populated with furniture and other objects. To get the overall performance of our policy, we report the success rate of search policy tested on randomly created scenarios. We further demonstrate the importance of enabling the head movement of the character for both better learning performance and better search behavior. Our work is a first step toward creating intelligent virtual agents with humanlike behaviors driven by onboard sensors, paving the road toward future robotic applications.

\section{Related Work} \label{sec:related}


Our research is inspired by prior work in visual navigation, deep reinforcement learning, and data-driven kinematic animation. Below we will review each of them in turn.

\subsection{Visual Navigation}

Training an autonomous agent to navigate complex environments from visual inputs has been an important topic in computer graphics, robotics, and machine learning~\cite{anderson2018evaluation,kempka2016vizdoom,mirowski2016learning,zhu2017target,pan2019zero}. 

Some of the work by Kuffner \etal \cite{kuffner1999perception} tackles the problem using path-planning and path-following algorithms that utilize the privileged information about the environment (e.g., floor layouts) and aims to generate collision free paths. Shim \etal \cite{shim2017automatic} and Wang \etal \cite{wang2018automatic} avoid the use of the privileged information and learn to approach the goal object which perform feature-based goal identification while tackling the searching via random exploration. In this work, the character is deprived of privileged information and perceives the world only via visual observations, which leads to a learned searching behaviour.

Enabling these agents to work with real images is essential for applying them to real-world applications. However, directly training visual navigation agents in real environments is expensive, especially since we usually want to train agents for a large variety of environments. To this end, researchers have developed simulated environments that leverages modern 3D scanning techniques to reproduce real-world scenarios and allow agents to observe photo-realistic visual inputs in a scalable way~\cite{xia2020interactive, habitat19iccv, ai2thor}. These tools enables rapid advancements in learning algorithms and neural network structures in training visual navigation agents~\cite{gupta2017cognitive, parisotto2017neural, zhang2017neural,fang2019scene, wijmans2020ddppo}. For example, Fang \etal \cite{fang2019scene} proposed a scene memory transformer architecture that saves history of observations into the memory and extracts relevant information using the Transformer architecture~\cite{vaswani2017attention}. Wijmans \etal developed a decentralized distributed proximal policy optimization (DD-PPO) that allows large scale training of visual navigation agents and demonstrated that with large scale training one can obtain agents that generalizes to novel scenarios~\cite{wijmans2020ddppo}.

Our method also utilizes the simulation tools developed by other researchers to train our virtual human in a realistic environment. Specifically, we used iGibson, which provides a suite of realistic indoor environments~\cite{xia2020interactive}. Unlike prior work in visual navigation that focus on agents moving in 2D or 2.5D spaces, e.g. mobile robots, our work solves visual navigation tasks with an additional challenge of controlling egocentric perspective.

\subsection{Deep Reinforcement Learning}

Deep reinforcement learning (DRL) provides a general framework for automatic design of controllers for complex motor skills from simple reward functions. Within the graphics community, researchers have applied DRL algorithms on a variety of physics-based control problems, such as locomotion~\cite{yu2018learning,peng2017deeploco,peng2018deepmimic}, manipulation~\cite{clegg2018learning}, aerial behaviors~\cite{won2017train}, and soft-body motion~\cite{min2019softcon}. However, most of these methods use low-dimensional character states or exploit privileged 3D information in the input space to the policy in order to simplify the learning problem. Directly learning a controller from egocentric vision inputs remains a challenging problem. Recent advancements in image-based deep reinforcement learning have shown promising progress in addressing this challenge~\cite{srinivas2020curl, he2019moco, oord2018representation}. For example, Srinivas \etal proposed to learn an embedding of the visual input by minimizing a contrastive loss between randomly cropped input images from the replay buffer~\cite{srinivas2020curl}. Their method demonstrated superior performance on a set of image-based robotic control problems. In our work, we apply the method by Srinivas \etal~\cite{srinivas2020curl} for training egocentric vision-based policies to accomplish the searching task.

Similar to our work, Merel \etal investigated the problem of creating full-body human motions with egocentric vision-based control policies~\cite{Merel:2019, Merel:2020}. In particular, they developed a hierarchical control scheme that exploits egocentric vision to coordinate low-level motor skill modules derived from motion capture demonstrations. Their learning approach is able to train the character locomotion and whole-body manipulation using visual inputs. Our work also takes 2D images as input, but our 3D scenes contain detailed geometry with photo-realistic appearance, resulting in a much more complex observation space than those used in the previous work. In addition, the vision inputs are more critical to the search task and require careful coordination between the locomotion and the gaze direction to enable the character to navigate in a cluttered environment while thoroughly surveying the environment.


\subsection{Data-driven Kinematic Animation}

Data-driven kinematic animation has been an effective approach for generating realistic human animations from example motion trajectories. Early work constructs graph-based structures to automatically transition between recorded motion clips ~\cite{lee2002interactive,arikan2002interactive,kovar2002motion}. Although these methods can successfully generate whole-body motions, the output motions are limited to motion clips in the database. To overcome this limitation, interpolation techniques based on linear bases~\cite{safonova2004synthesizing,chai2005performance} or statistical transition models~\cite{chai2007constraint,lee2010motion} are adopted for predicting more expressive motions from a smaller number of examples. These methods are further extended by exploiting neural networks, such as conditional Restricted Boltzmann Machine (cRBM)~\cite{taylor2009factored} or an Encoder-Recurrent-Decoder (ERD) network~\cite{fragkiadaki2015recurrent}. Recently, many researchers have demonstrated that deep neural network can successfully learn human motion manifolds for bipedal locomotion~\cite{holden2016deep,holden2017phase}, quadrupedal locomotion~\cite{zhang2018mode}, object interactions~\cite{starke2019neural}, and motion retargetting~\cite{aberman2019learning}. We utilize previous work by Holden \etal \cite{holden2017phase} and Starke \etal \cite{starke2019neural} to generate detailed full-body animations from an abstract trajectory planned by the search policy.
\section{Overview} \label{sec:overview}

\graphicspath{{figures/}}


Given a realistic 3D indoor scene populated with furniture and household items, we develop a virtual human capable of using its own egocentric vision and locomotion capability to search for any randomly placed target object in the scene, including those occluded by furniture or due to the layout of the room. We take a hierarchical approach which consists of two components, a \emph{search policy} operates on an abstract agent to determine the movement and gaze direction at every time step, and a \emph{motion synthesis} module that synthesizes the kinematic motion on a full-body character to realize the actions determined by the search policy.

We made a few assumptions in our framework. The vision input includes RGBD images and a mask channel in which the target object has been segmented and labeled. Automatic segmentation and object recognition for general scenes and objects from raw images is a challenging computer vision problem, beyond the scope of this work. In addition, we assume that the 3D environment has furniture and partial divisions that block the line of sight, but it is roughly one connected space, as we do not attempt to solve a maze navigation problem.

\section{Search Control Policy}
While end-to-end DRL approaches have demonstrated success in learning motor skills, solving a visuomotor policy with a detailed full-body character in a large and highly textured environment remains challenging due to the large and complex observation space of the agent. In this work, we propose to first use the learning approach for training an agent-agnostic search policy that has a vision-based observation space but an abstract action space. Once trained, the search policy can be applied to characters with various kinematics to synthesize search behaviors with full-body movements. 
To this end, we define an abstract model that consists of a cylinder-shape main body and a camera connected to the main body via a universal joint with two degrees of freedom (dofs). The abstract model can move around in the 3D space while the camera can point at different directions independently from the body movement. This additional head movement allows a character to simultaneously navigate and look around, which is essential to model human-like search behaviors.

\subsection{Problem Formulation}
We formulate the vision-based search task as Partially Observable Markov Decision Processes (POMDPs), $(\mathcal{S}, \mathcal{O}, \mathcal{A}, \mathcal{T}, r, p_0, \gamma)$, where $\mathcal{S}$ is the state space, $\mathcal{O}$ is the observation space, $\mathcal{A}$ is the action space, $\mathcal{T}$ is the transition function, $r$ is the reward function, $p_0$ is the initial state distribution and $\gamma$ is a discount factor. We take the approach of model-free reinforcement learning to find a policy $\pi$, such that it maximizes the accumulated reward:
\begin{equation}
\label{eq:rl}
    J(\pi) = \mathbb{E}_{\mathbf{s}_0, \mathbf{a}_0, \dots, \mathbf{s}_T} \sum_{t=0}^{T} \gamma^t r(\mathbf{s}_t, \mathbf{a}_t),
\end{equation}
 where $\mathbf{s}_0 \sim p_0$, $\mathbf{a}_t \sim \pi(\mathbf{o}_t)$, $\vc{o}_t \sim c(\vc{s}_t)$ and $\mathbf{s}_{t+1}=\mathcal{T}(\mathbf{s}_t, \mathbf{a}_t)$. The state space contains 3D information of the environment and the global position and orientation of the agent for reward evaluation during training, but it is not available to the policy during testing. Instead, the policy can only access limited information observable to the onboard sensors. Our POMDP is defined as follows: \\

\begin{figure}
    \centering
    \includegraphics[width=\linewidth]{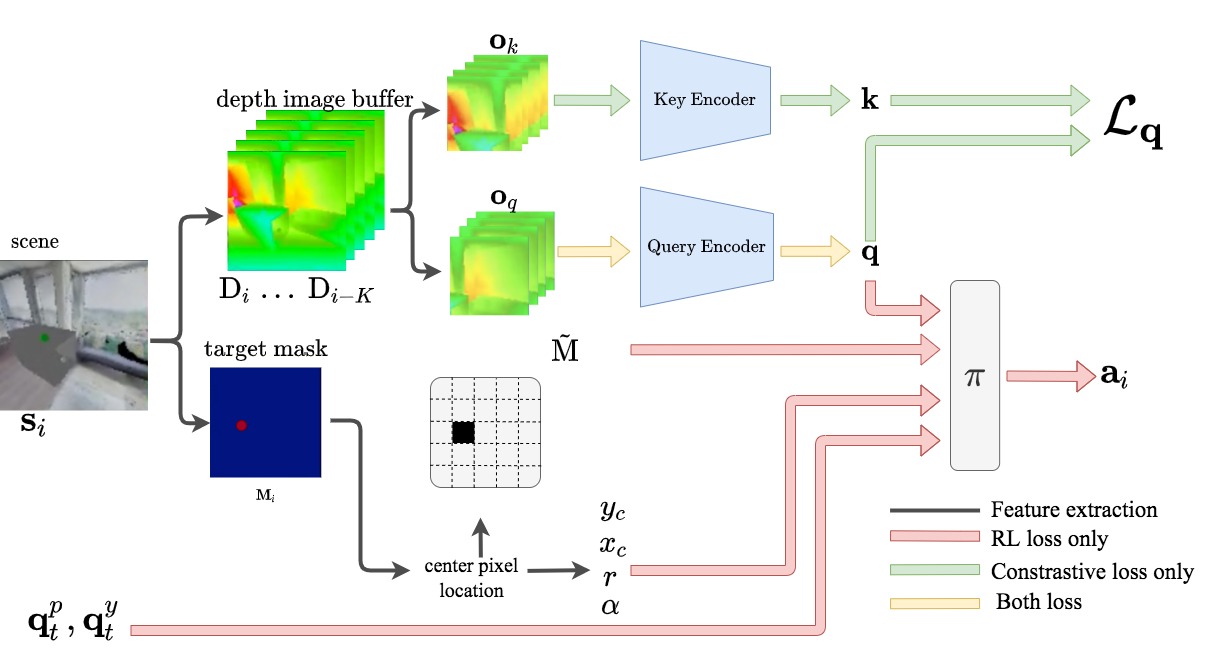}
    \vspace{-5mm}
    \caption{Overview of the learning pipeline for training the search policy.}
    \vspace{-5mm}
    \label{fig:observations}
\end{figure}

\noindent \textbf{Observation space.} The observation space consists of two sensing modalities: vision and proprioception (Figure \ref{fig:observations}). The proprioception for the abstract model only contains the joint position between the camera and the cylinder: the pitch angle $q^{p}$ and the yaw angle $q^{y}$. The agent does not have an access to its global position and orientation. 

The vision perception is represented as 2D images observable by standard RGBD cameras, augmented by mask images that provide the segmentation of the target object. For the searching tasks, we exclude the color information (RGB) because we found that the depth and mask images alone contain sufficient information for finding collision-free paths to complete the task when navigating in a cluttered 3D scene. The depth image $\mat{D}$ ($84 \times 84$), obtained with the field of view (FOV) angle of $90^{\circ}$, has the maximal depth at $5$~m, normalized to the range of $[0, 1]$. This setting has been proven effective for visual navigation tasks, such as Point-Goal navigation~\cite{Savva_2019_ICCV}.

The mask image $\mat{M}$ is a binary image which contains $1$s at the target object's pixel locations and $0$s otherwise. When the target object is not in the field of view, $\mat{M}$ provides no information. We process the raw mask image $\mat{M}$ to obtain a feature vector $\vc{m} = [x_c, y_c, r, \alpha, \tilde{\mat{M}}]$, where $x_c$ and $y_c$ are the average coordinates of the pixels with value $1$, $r=\sqrt{x_c^2 + y_c^2}$ and $\alpha=arctan(y_c, x_c)$ are their polar coordinates, and $\tilde{\mat{M}}$ is the downsampled mask image of size $5 \times 5$. Using our mask feature vector instead of the raw mask image reduces the dimensionality of the state space and forces the agent to learn a policy agnostic to the object shapes.
Note that $x_c, y_c, r$ and $\alpha$ are all defined in the image frame. 
The observation at every time step $t$ is defined as $\vc{o}_t = [\vc{D}_t, \vc{D}_{t-1},$ $\cdots, \vc{D}_{t-K+1}, \vc{m}_t, q^{p}_t, q^{y}_t]$ where we concatenate the $K$ history of the depth images so the policy has some ``short-term memory'' of the environment. We set $K=5$ for all the experiments.

\noindent \textbf{Action space.} We define a compact action space that only determines the agent's 2D global movement and the camera direction. Specifically, the action vector is defined as $\vc{a} = [\Delta x, \Delta y,$ $\Delta \theta, \Delta q^{p}, \Delta q^{y}]$, 

which are the relative movements in the forward direction, lateral direction, yaw orientation, camera pitch angle, and camera yaw angle, respectively.
We use the action at the current time step to modify the target global configuration and camera angles for the next time step. The next state of the abstract model is simulated by tracking the target global configuration and the camera pose using position control in a physics simulator: $\vc{s}_{t+1} = \mathcal{T}(\vc{s}_t, \vc{a})$.

\noindent \textbf{Reward function.} Unlike policy execution, when we evaluate the reward function during training, we utilize all 3D information relevant to the reward calculation, such as global position and orientation of the agent, the 3D coordinate of the target object and the 3D meshes of the environment. Inspired by the work of Savva \etal \cite{Savva_2019_ICCV}, we define the following reward function:
\begin{equation}
    r(\vc{s}_t) = w_1 r_s(\vc{s}_t) + w_2 r_d(\vc{s}_t) + w_3 r_l(\vc{s}_t) + w_4 r_c(\vc{s}_t).
\end{equation}
$r_s$ is the success reward of $10$ when the agent successfully finds the object, which is only awarded once the success-checking terminal condition is invoked.
$r_d$ measures the negative distance to the goal while returns $0$ if the goal is not visible, which encourages the agent to move toward and look at the target. 
$r_l$ is the live penalty of the value $-0.1$. 
Finally, $r_c$ checks the collision between the cylinder (the main body) and the environment mesh and penalizes collision by $r_c(\vc{s}_t) = clip(-0.1 n_{col}, -3.0, 0.0)$, where $n_{col}$ is the number of collision in the history. 
For all experiments, we use the same weight vector $[1.0,1.0,1.0,0.1]$.

\noindent \textbf{Initial state distribution.} For training a robust searching behavior, we randomize both the agent's initial location and the target object's location at each episode. We collect the candidate locations for both initial positions by sampling random places and filter out the invalid candidates that collide with any other objects. Note that the target object is not necessarily always blocked from the agent's line of sight and might be visible immediately, which successfully facilitates the learning process.

\noindent \textbf{Termination conditions.} There are two occasions at which the trajectory rollout can be terminated. First is the time based condition, which caps the total number of actions in the environment at $T_{max}$ steps, which is set to $100$ for all the experiments. 
Second condition control is a success check, which is triggered once the object is in the nearby proximity of $0.5m$ and the agent is observing the object, i.e. mask image $\mat{M}$ contains some 1s. 

\noindent \textbf{Domain randomization.} Similar to many sim-to-real transfer learning problems, the success of the search policy depends on whether it can be transferred to the target character with different state and action space. We apply the approach of domain randomization to increase robustness of the search policy when transferred to a different character. We identify that the global vertical position of the character can be drastically different from that of the abstract model due to its designed body height. We therefore randomize the height of the abstract model during training for the duration of a trajectory rollout in the range of $[1.0, 1.8]m$. In addition, legged characters may exhibit natural vertical oscillation during locomotion, while the abstract model moves at a constant height. Therefore, we inject white noises with the range of $[-0.1, 0.1]m$ to the vertical position of the abstract model at each time step. The randomized vertical movement will affect visual observations by changing the camera position, essential to success transfer of search policy to different characters.

\subsection{Policy Training Process}
Training control policies for simulated character with visual perception input has several challenges. First, these policies usually have complex structures and a large number of parameters, making it computationally expensive to obtain reliable gradients for updating the policy. In addition, learning to extract useful features from images depending solely on the task reward might lead to sub-optimal features that do not generalize well to new scenarios.

We train the policy using Soft Actor Critic (SAC) and Contrastive Unsupervised Representations for Reinforcement Learning (CURL), as proposed in the work of Srinivas \etal~\cite{srinivas2020curl}.
SAC~\cite{haarnoja2018soft} is an off-policy model-free reinforcement learning algorithm, which has been applied to challenging robotic control problems with desirable sample efficiency~\cite{haarnoja2018learning}.
CURL augments the SAC algorithm for learning effective features from high dimensional pixels by jointly optimizing the DRL loss and a contrastive loss to learn a compact latent space.

Specifically, from each input image, CURL will randomly crop two sets of smaller images named queries and keys. These cropped images are then passed through a query encoder and a key encoder to obtain a low-dimensional latent representation of the image $\mathbf{q}$ and $\mathbf{k}$. CURL formulates a constrastive loss:
\begin{equation}
    \mathcal{L}_{\vc{q}}=\log \frac{\exp \left(\vc{q}^{T} W \vc{k}_{+}\right)}{\exp \left(\vc{q}^{T} W \vc{k}_{+}\right)+\sum_{i=0}^{K-1} \exp \left(\vc{q}^{T} W \vc{k}_{i}\right)},
\end{equation}
where $W$ is a learn-able weight matrix and $\mathbf{k}_+$ are keys that are from the same time instance of $\vc{q}$. The constrastive loss encourages the encoded latent features of queries and keys from the same time instance to be close to each other while being far away from the latent features from different time instances under a bilinear product distance. We optimize this contrastive loss jointly with the RL objective (Equation \ref{eq:rl}). Figure \ref{fig:observations} illustrates the data flow of our learning algorithm and indicates the paths where the gradient information is propagated through. We refer readers to the original paper of SAC and CURL for more details about the algorithms. 

We represent our search policy as a convolutional neural network. The history of depth image is passed through a Pixel CNN into an embedding of 128 dimensions \cite{srinivas2020curl}. The depth image embedding is then concatenated with the mask feature and the agent state, which are then fed into 3 fully connected layers with 1024 neurons to obtain the final action output. We train the search policy using SAC with CURL for $0.75$ million simulation samples.

\section{Full-body Motion Synthesis} \label{sec:full-body}

The search policy generates global/root configuration and head/camera movements for the abstract model. However, this trajectory cannot be transferred directly to the actual character due to the discrepancy in the state and action spaces and the transition functions, between the actual character and the abstract model. In particular, the policy will receive different input images due to the oscillating head height of the characters and due to the character exhibiting the walking motions and not searching. Furthermore, given the same command, such as moving forward, the characters will achieve different resulting states depending on the transition function of the locomotion model. The differences in the height of the characters can be mitigated by domain randomizing and injecting noise to the height of the abstract model during training to robustify the controller. While, the difference in the head motion will be resolved by querying the controller as if it was to follow the trajectory generated by actual character.

To overcome the discrepancy in the transition function of locomotion model, we employ an online replanning scheme to generate character motions that best match the planned trajectory from the abstract model. Our method is analogous to the model predictive control (MPC) framework in that every time step we replan a long trajectory using the abstract model and execute only a small portion of the trajectory under full-body dynamics.
Please note that the trajectories must be collision-free for both abstract and full-body dynamics.

At current time step during testing, we first rollout the search policy (with abstract model) for $T$ time steps from the current state and observation, $\vc{s}_0$ and $\vc{o}_0$, by repeatedly querying the policy, $\vc{a}_t = \pi(\vc{o}_t)$, stepping forward in the environment, $\vc{s}_{t+1} = \mathcal{T}(\vc{s}_t, \vc{a}_t)$, and observing the environment, $\vc{o}_{t+1} = \mathrm{render}(\vc{s}_{t+1})$, \ie rendering the environment to images. Note that the state of the abstract model $\vc{s}$ includes the root configuration and the head pose. We pass the planned state trajectory $\vc{s}_0, \cdots \vc{s}_T$ to the locomotion generator Phase-Functioned Neural Network (PFNN)~\cite{holden2017phase} or Neural State Machine (NSM)~\cite{starke2019neural} which generates legged motion on a human character to match the plan. Our method is agnostic to the full-body motion generator if it can synthesize a reasonable full-body motion trajectory for the given abstract plan: we will generally refer them to as ``MG'' thereafter. 

However, the root and body/head states along the full-body trajectory are likely to deviate from the planned state trajectory due to the difference in the transition function between MG and our abstract model. Applying the strategy of online replanning, we only consider the first $M$ (where $M \ll T$) steps of the full-body trajectory $\vc{q}_0, \cdots, \vc{q}_{M-1}$ to be valid, and replan the abstract model at the $M^{th}$ step.

One issue with our online replanning scheme is that the locomotion generator, MG, does not generate vision perception during motion synthesis, but replaning at the $M^{th}$ step requires a short history of depth images as the short-term memory. Furthermore, the planned head motion becomes suboptimal for the searching task since the traversed trajectory deviates from the plan. As such, the vision observations generated by the head poses in $\vc{q}_0, \cdots, \vc{q}_{M-1}$ can also result in suboptimal next plan. To overcome this issue, we iteratively update the history of abstract model's state $\vc{s}_t$ from $t=0$ to $t=M-1$ using the root configuration from the full-body pose $\vc{q}_t$ and the head pose from the search policy. We re-generate observations $\vc{o}_t$ using the modified abstract state and store the depth images in the buffer to recover the ``memory lapse'' from $t=0$ to $t=M-1$. Finally, the abstract model makes a new plan from $t=M$ to $t=M+T$ with optimal head movements and restored memory (history of depth images). Our algorithm applied at every $M$ time steps can be summarized in Algorithm 1.

\begin{algorithm}
\caption{Online Motion Replanning and Synthesis}\label{alg:motion_synthesis}
\begin{algorithmic}[1]
\State Input: $\vc{s}_0, \vc{o}_0$, $\mathbf{q}_0$
\State $\mathcal{B} = \{\vc{s}_0\}$ \Comment{initialize plan buffer}
\For{$t=0:T-1$} \Comment{generate an abstract trajectory}
\State $\vc{a} = \pi(\vc{o}_t)$ \Comment{query policy}
\State $\vc{s}_{t+1} = \mathcal{T}(\vc{s}_t, \vc{a})$ \Comment{advance in environment}
\State $\vc{o}_{t+1} =$ render($\vc{s}_{t+1}$) \Comment{render observation}
\State Store $\vc{s}_{t+1}$ in $\mathcal{B}$
\EndFor

\State $\vc{q}_0, \cdots, \vc{q}_{M-1} = \mathrm{MG}(\mathcal{B})$ \Comment{generate $M$ human motion poses} 

\For{$i=0:M-1$} \Comment{regenerate head orientations \& history}
\State $\vc{s}_i.\mathrm{root} = \vc{q}_i.\mathrm{root}$ 
\Comment{update the root position}
\State $\vc{s}_i = \mathrm{set\_root\_height}(\vc{q}_i)$ 
\Comment{update the camera height}
\State $\vc{o}_i =$ render($\vc{s}_i$) 
\State Update depth image buffer with $\vc{o}_i$
\State $\vc{a} = \pi(\vc{o}_i)$ \Comment{query policy with new observation}
\State $\vc{s}_{i+1} = \mathcal{T}(\vc{s}_i, \vc{a})$ \Comment{modify camera pose according to $\pi$}
\State $\vc{q}_{i+1}.\mathrm{head} = \vc{s}_{i+1}.\mathrm{head}$ \Comment{overwrite head orientation }
\EndFor
\State $\vc{o}_{M} = \mathrm{render}(\vc{s}_M)$ \Comment{render at last camera pose}
\State Return $\vc{s}_M, \vc{o}_M, \vc{q}_0, \cdots, \vc{q}_{M-1}$
\end{algorithmic}
\end{algorithm}

\section{Evaluation}
\begin{figure}[ht!]
    \centering
    \includegraphics[width=0.99\linewidth]{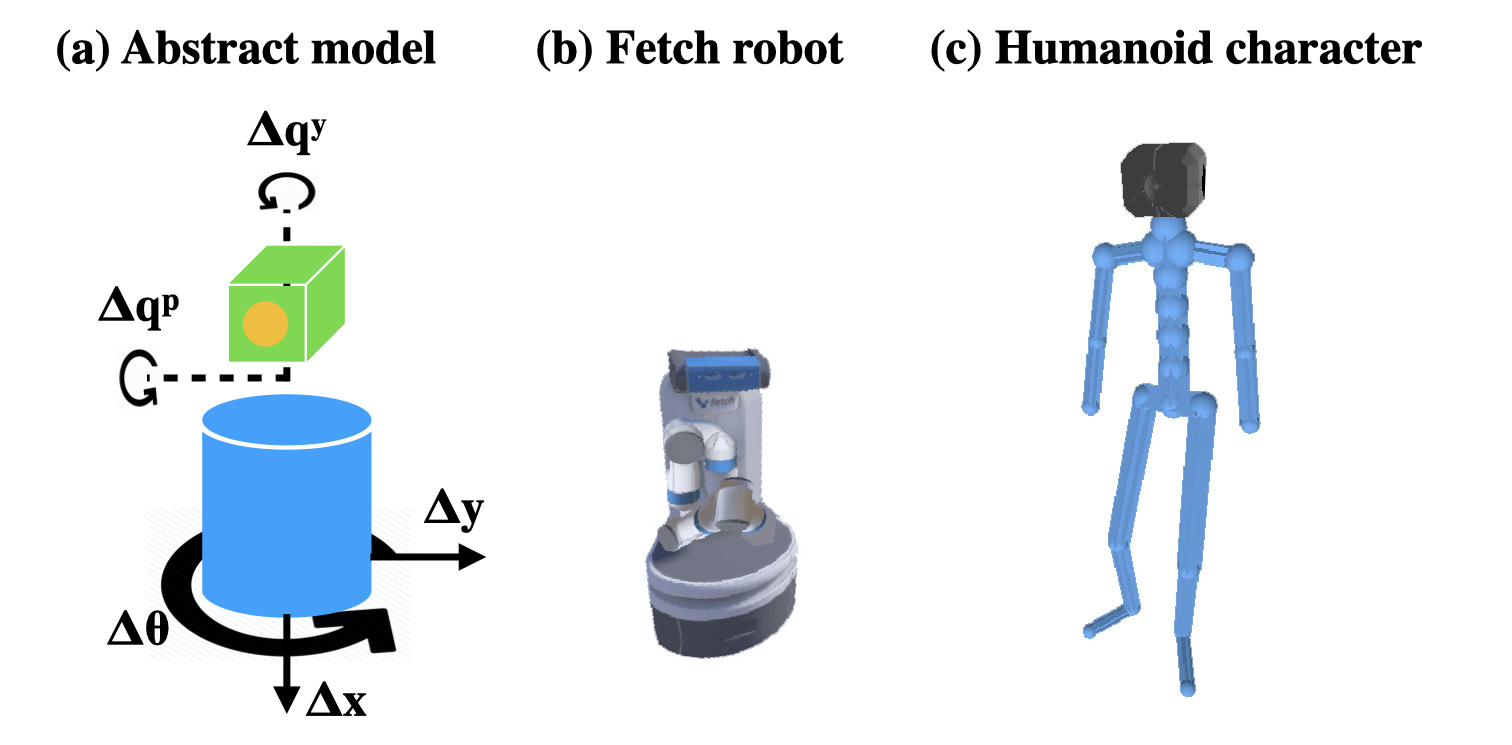}
    \caption{An abstract model (left) for policy search and two animated characters,the Fetch robot (middle) and a humanoid character (right), for full-body animation.}
    \label{fig:characters}
\end{figure}

\begin{figure}[ht!]
    \centering
    \includegraphics[width=\linewidth]{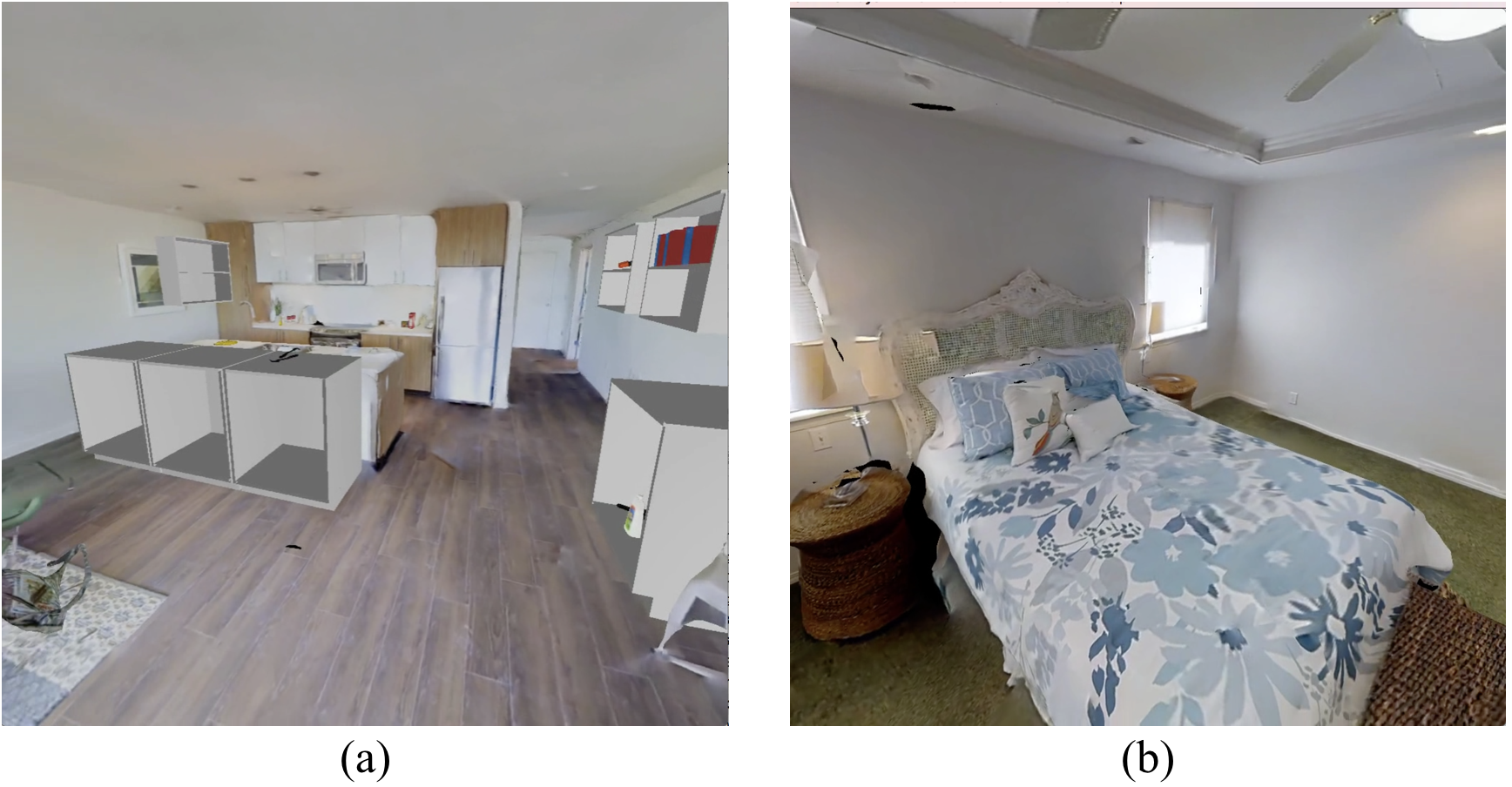}
    \vspace{-6mm}
    \caption{The home scene used in our experiments. (a) a kitchen and a living area separated by a countertop. (b) a bedroom connected to a bathroom. }
    \label{fig:scene_example}
\end{figure}

We evaluate our method on a humanoid character and a wheel-based robot character, Fetch Robot \cite{wise2016fetch}. For the humanoid character, we applied two motion generators to synthesize the full-body motion. Figure \ref{fig:characters} shows the humanoid character and the robot we use in our experiments. We test our search controller in a modern home scene with an open kitchen and a living area separated by a countertop, as well as a master bedroom connected to a bathroom (Figure \ref{fig:scene_example}). To increase complexity of the scene, we add a few open cabinets in which the target objects can be placed. We use iGibson ~\cite{xia2020interactive} environment that provides 3D scans reconstructed from realistic indoor environments and a photorealistic renderer to generate vision inputs to the character. For physics simulation, we use PyBullet \cite{pybullet} to simulate the motion of abstract model and to check collision with the environment.

\begin{figure}[ht!]
    \centering
    \includegraphics[width=\linewidth]{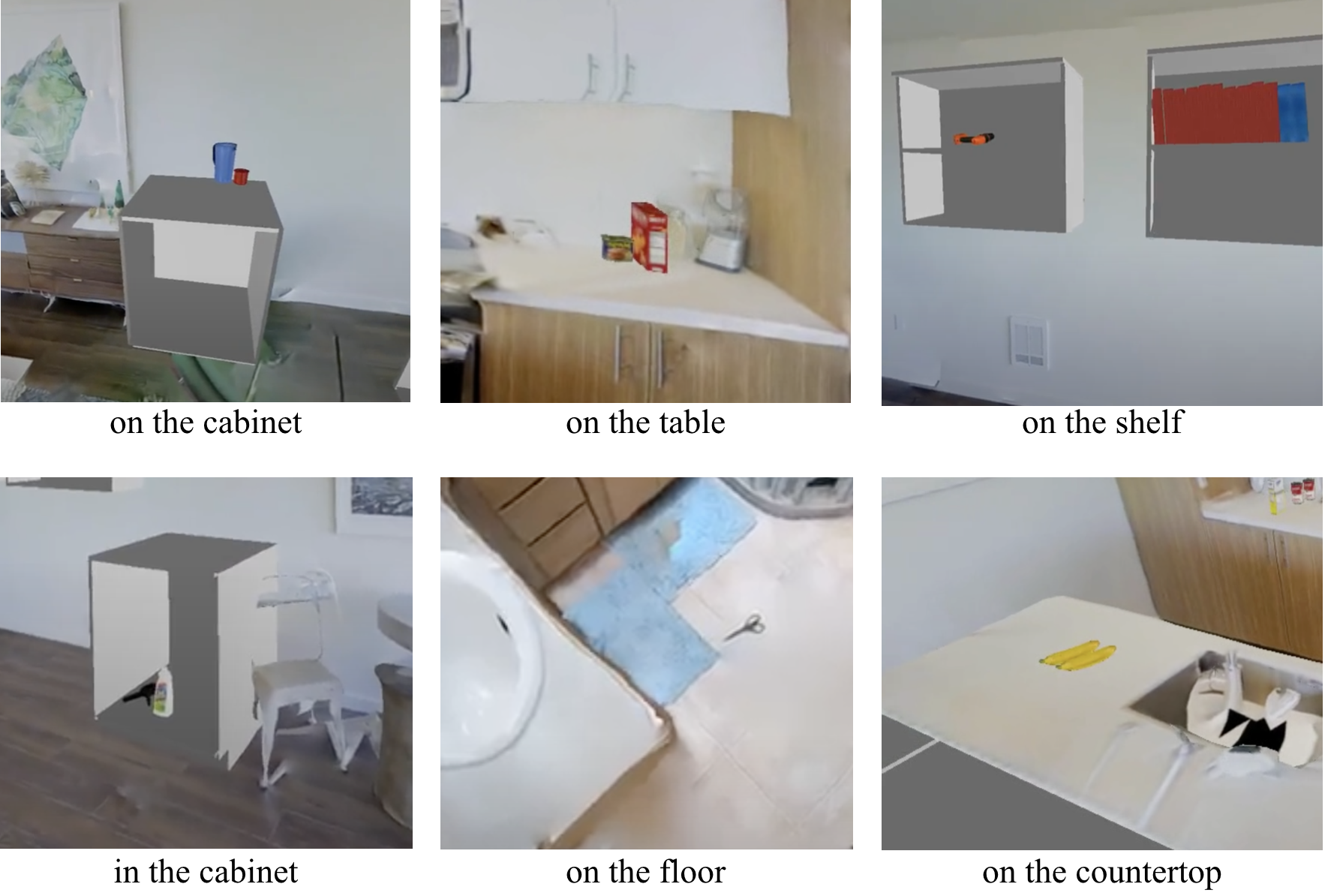}
    \caption{Examples of object placements in our experiments.}
    \vspace{-3mm}
    \label{fig:placement_illus}
\end{figure}

\subsection{Evaluation of Search Policy}
We generate $100$ random scenarios to evaluate the success rate of the search policy for an abstract model. At the beginning of each test, the agent is randomly assigned to a collision-free location in the scene with random orientation. Similarly, the target object is placed randomly on any surface in the scene, including the interior of cabinets~(Figure \ref{fig:placement_illus}). 
If the agent can get close to the target object within 100 steps ($\sim$15 seconds),
we consider it a successful trial. 

Since the search policy will be used by different agents with specific body heights, we evaluate the performance of the search policy by setting the abstract model to three different heights: 1.65m, 1.05m, and 0.45m, where the first two correspond to the height of the human character and the Fetch robot, and 0.45m is selected for further comparison.
We compute the success rate of the policy with those three different heights and show the results in Table~\ref{table:search_performance}. In general, the advantage of height gives the tall characters a better view of surfaces while shorter characters struggle to see objects placed on the surface above their heights. As such, there is a near 20\% drop in success rate for the shortest character. 

We also test the policy on two sampling schemes of the target locations of objects: 1) sampling everywhere except for the inside area of the low cabinets on the floor, and 2) sampling everywhere. 
The results show that the tall characters (1.05m and 1.65m) perform worse on those challenging cases where objects are inside the low cabinets, while the success rate of the shorter character (0.45m) increases when we allow objects to be sampled inside the cabinet.

\begin{table}[ht]
\begin{center}
\begin{tabular}{|l|c|c|}
\hline
\textbf{Head height}  &  \textbf{Excluding cabinets}  & \textbf{Everywhere} \\
\hline
0.45m  &  50\%  & 58\% \\
\hline
1.05m & 92\% & 75\%\\
\hline
1.65m & 90\% & 76\%\\
\hline
\end{tabular}
\caption{Performance of the search policy with different camera heights and object locations. In general, abstract models with higher camera locations show better success rates. However, lower cameras are beneficial if we include the challenging case of the object is hidden in the cabinets on the floor.}

\label{table:search_performance}
\end{center}
\end{table}

\subsection{Comparison to Search Policy without Head Movements}

\begin{figure}[ht!]
\centering
  \centering
  \includegraphics[width=0.9\linewidth]{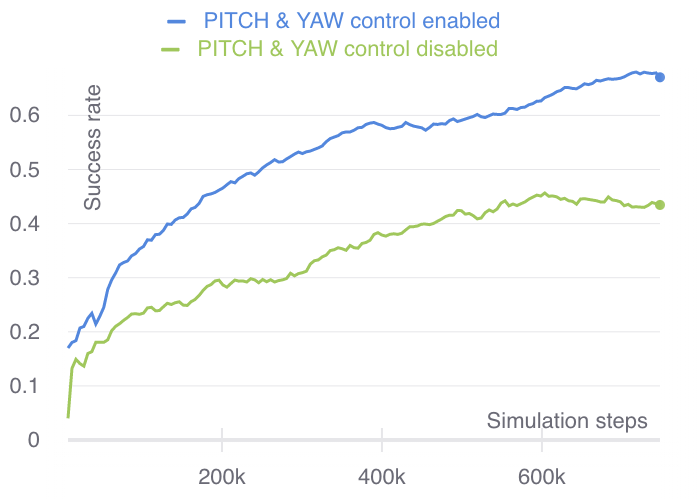}
  \captionof{figure}{Training curves comparison of abstract agents with and without head movements. The result indicates that the agent with head movements shows a 20\% higher success rate than the agent without head movements.
  }
  \label{fig:training_curve_withouthead}
\end{figure}

Our key hypothesis is that active head movements lead to more effective search behavior by allowing the character to look at different parts of the scene.
To evaluate this, we train a baseline search policy for an abstract character without the degrees of freedom to move its head relative to the body. To look around the environment, the agent needs to rotate its entire body around. The result shows that the head movement is crucial to achieve 20\% higher success rate in the searching task (Figure~\ref{fig:training_curve_withouthead}).

\subsection{Evaluation of Full-body Characters}

To evaluate the performance of our algorithm we use human character with the height of 1.65m. To animate the character as a kinematic motion generation model we use two distinct Motion Generators: Phase-Functioned Neural Network (PFNN)~\cite{holden2017phase} and Neural State Machine (NSM)~\cite{starke2019neural}. We also utilize both to generate legged locomotion for the character and incorporate the discrepancy using the online replanning control scheme described in Section \ref{sec:full-body}. 
Additionally, we apply the searching policy on another drastically different model, a Fetch Robot with height of 1.05m. Fetch is a wheel-based robot with telescopic degree of freedom to adjust its height. Due to similarity between our control model and Fetch's differential drive we directly apply the actions produced by the search policy on Fetch.

We show that the search policy can be successfully realized by the full body characters even for challenging cases in which the target objects are placed inside of cabinets, on the other side of room, or occluded by furniture. Different search strategies emerge around different locations in the scene. For example, the character tends to move slower around cabinets and look carefully at the interior part where objects are likely to be placed. The character also utilizes backward steps to have a better view of the surface in front of it. The results are best viewed in the supplemental video where we show the full-body motion in the first-person-view as well as multiple third-person views.

\subsection{Performance Analysis }
To further analyze the performance of our method and understand the choices of different hyperparameters, we construct a few baselines for comparison:


\begin{itemize}
  \item \textbf{1-step:} To evaluate the importance of long-term planning for the searching task, we design a baseline to control the agent by querying the searching policy for one single action and passing it to MG. This baseline does not utilize the long term trajectory planning but instead extrapolates a straight line path in the action direction, with respective orientations. We evaluate two settings with longer and shorter horizons for MG.
  \item \textbf{Noisy Search:} To evaluate the importance of the searching policy before the target object is seen, we create a simple control scheme which in the presence of the object will query the trained searching policy (to use it as an approach mechanism with obstacle avoidance), and when there is no object apply actions sample uniformly at random to move and look around the scene.
  \item \textbf{Ours with different Ms:} To understand the effects of the hyperparameter choices of motion synthesis algorithm, we evaluate multiple values controlling the parameter $M$ which specifies the number of steps MG returns when attempting to follow the trajectory.
\end{itemize}

The above baselines and our method are compared using the following metrics:
\begin{itemize}
    \item \textbf{Number of attempts:} Number of trajectories generated by the abstract searching policy before the character reached the goal.
    \item \textbf{SPL:} Success weighted by the path length, represented as:
    $$\frac{1}{N} \sum_{i=1}^{N} S_{i} \frac{\ell_{i}}{\max \left(p_{i}, \ell_{i}\right)},$$
    where $\ell_{i}$ is the shortest distance path length, $p_{i}$ is a traversed path length, and $S_i$ is a binary indicator if the rollout success.
\end{itemize}

\begin{figure*}
    \centering
    \vspace{-2mm}
    \includegraphics[width=0.43\linewidth]{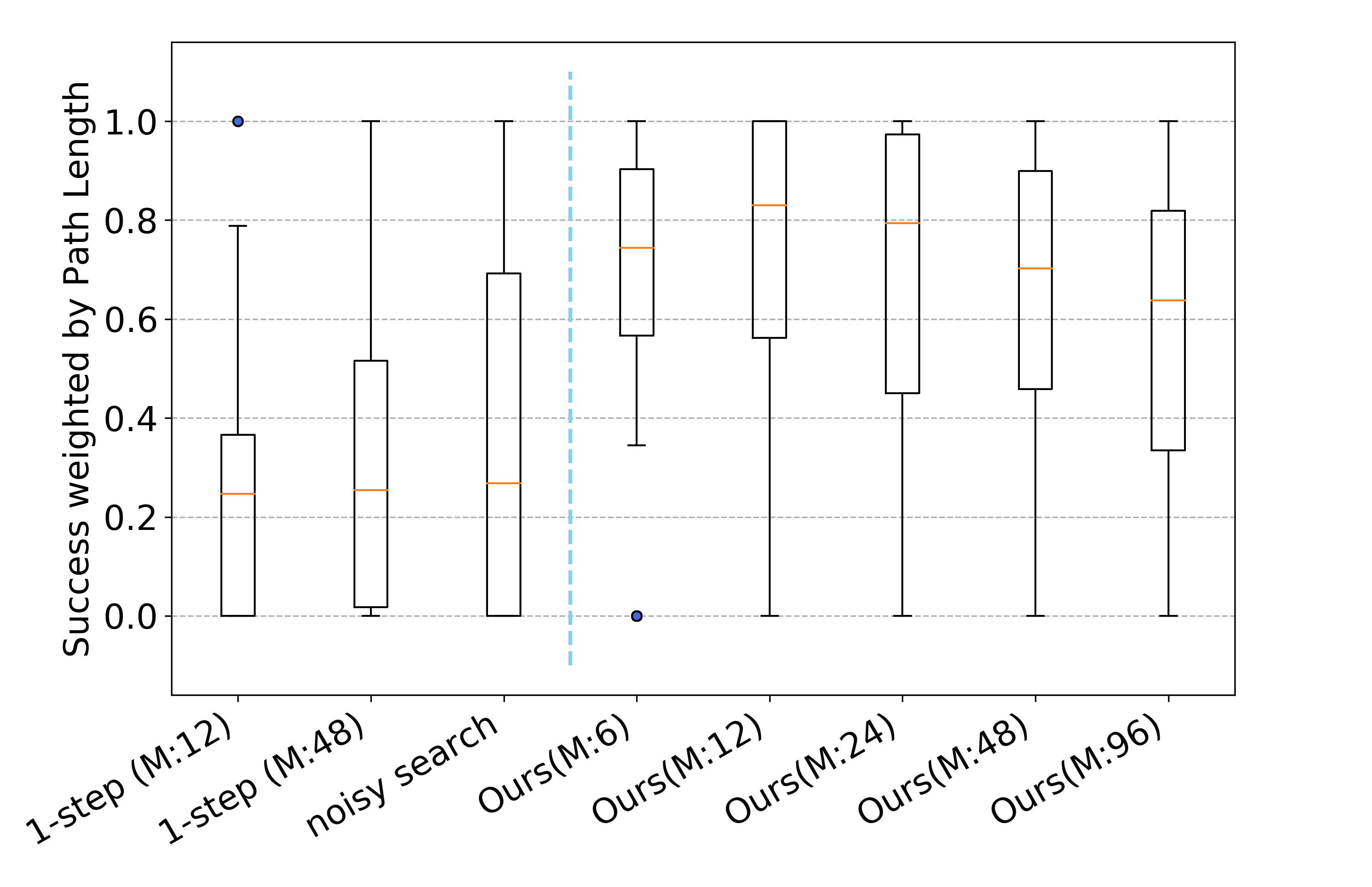}
    \vspace{-3mm}
    \includegraphics[width=0.43\linewidth]{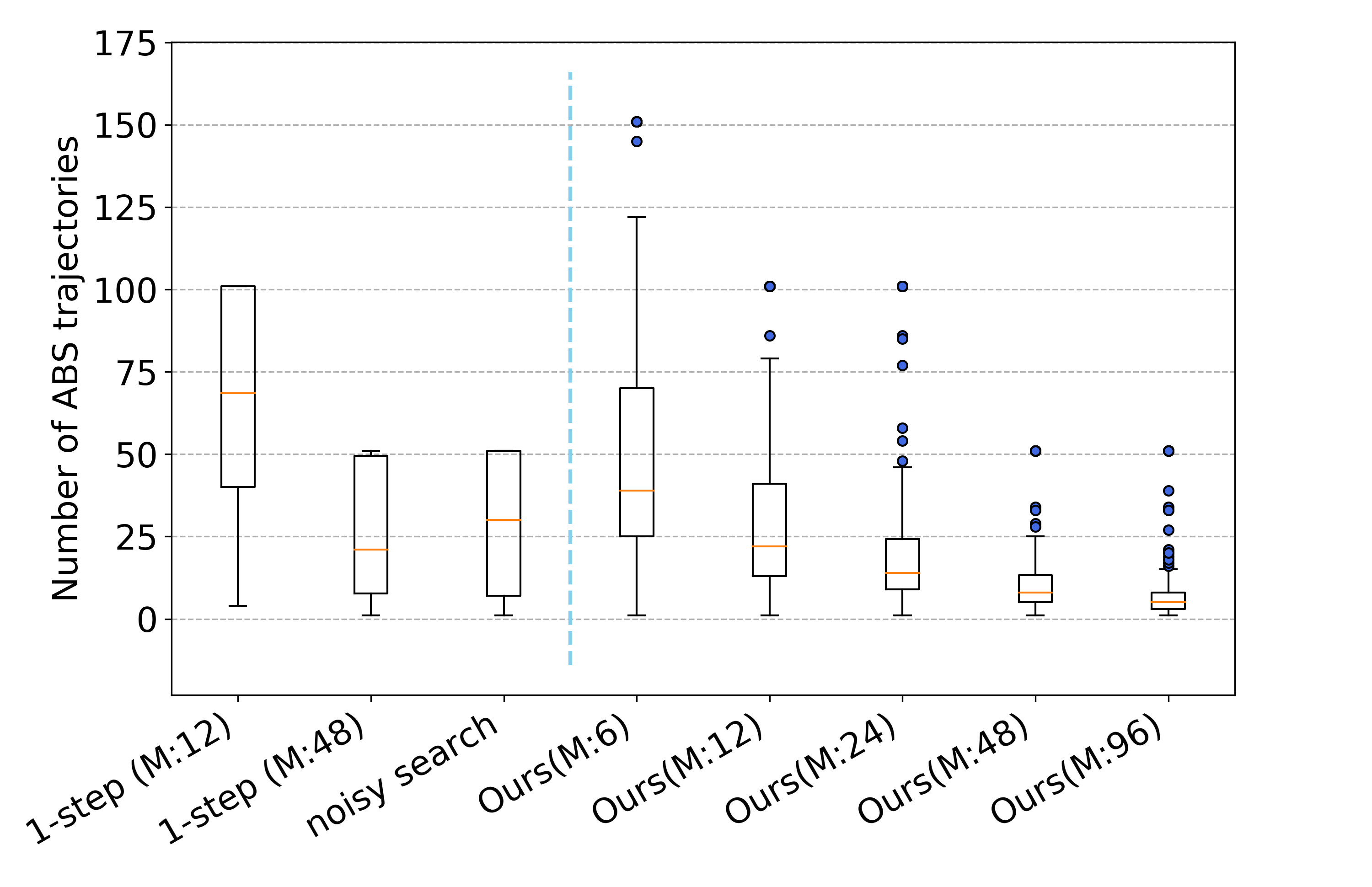} 
    \caption{
    Performance of baselines (to the left of dashed line) and our online motion planning with different hyperparameter choices (to the right of dashed line).
    \textbf{left} - Success weighted by the path length - showing the importance of long-term trajectory planning and that simple noise based exploration is not sufficient for finding the object in the designed environment.
    \textbf{right} - shows that a larger number of abstract trajectories generated by abstract model is required for smaller values of $M$, which is a trade-off for higher SPL.
    \vspace{-3mm}
    }
    \label{fig:performance}
\end{figure*}

Figure \ref{fig:performance} left shows that \textbf{1-step} performs poorly for both MG horizons compared to our method with long-term planning. On average, our method performs 40\% hire than \textbf{1-step} on SPL metric. 
The performance of \textbf{noisy search} is also worse than our method by 40\%, which shows that structured exploration is required to efficiently find the target object. 
The performance of our method is sensitive to the choice of the hyperparameter $M$ (the number of executed frames from the motion plan).
In our experiments, smaller values of $M$ perform better in terms of the SPL because the algorithm replans more frequently, but it also requires more computational costs to generate a larger number of generated abstract trajectories (Figure \ref{fig:performance} right).


Most common failure modes for the baselines are related to collisions with the environment obstacles, which causes the character to get stuck in the furniture or the walls. 
Collisions can also happen to the full-body character if it significantly deviates from the plan.
However, we found that collisions of the full-body character occur less frequently because the agent is trained with penalty to avoid the obstacles. Some example scenarios are illustrated in Figure \ref{fig:failures}.

\vspace{-3mm}

\section{Discussion and Conclusion}

\begin{figure}
    \centering
    \includegraphics[width=0.85\linewidth]{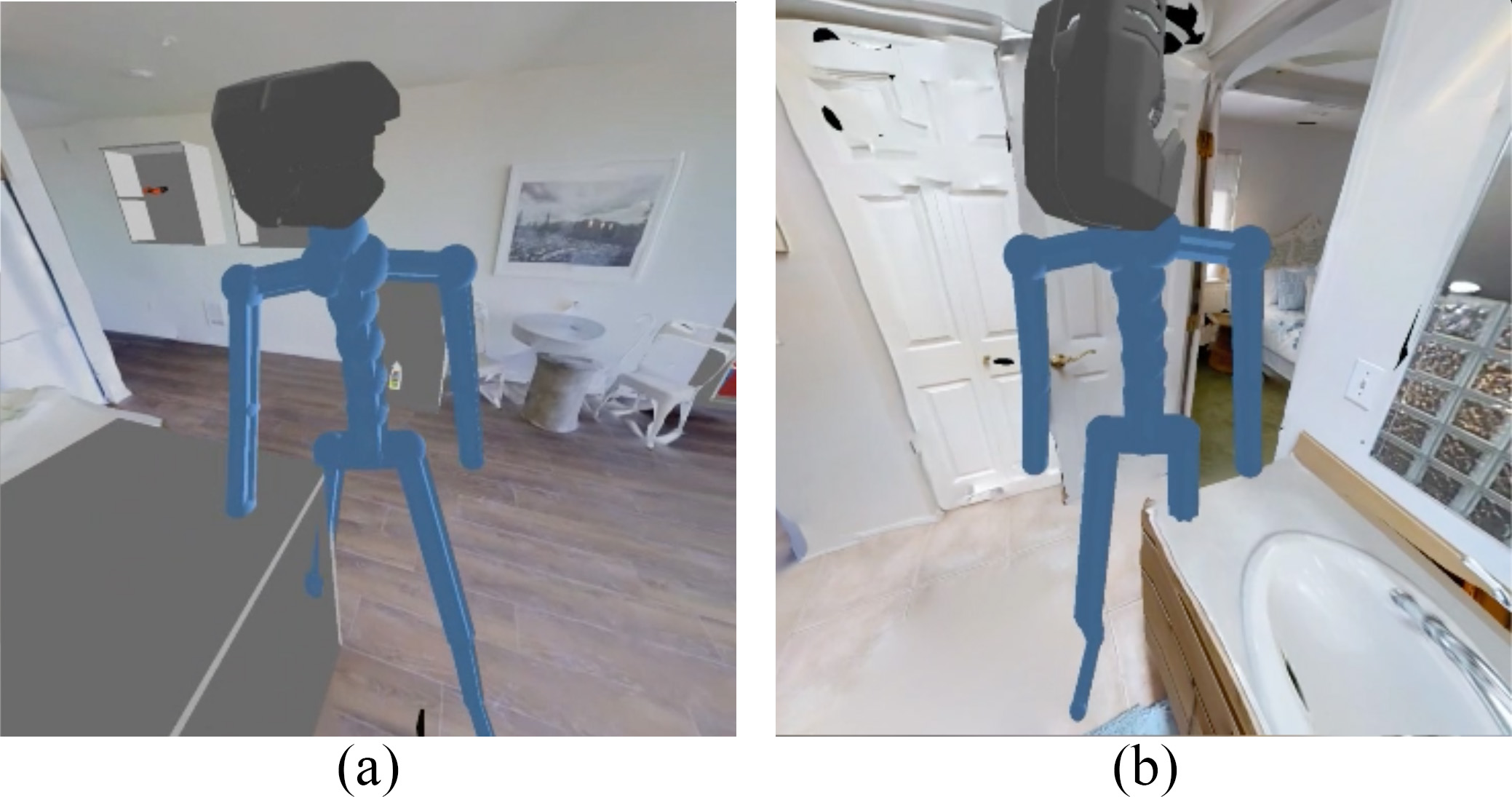}
    \vspace{-3mm}
    \caption{Failure cases where the character penetrates furniture. }
    \vspace{-5mm}
    \label{fig:failures}
\end{figure}

In this work, we have developed a virtual character exhibiting searching behaviors inspired by the human. We have demonstrated that by training a agent-agnostic search policy and using a replanning algorithm for transferring the planned abstract motion to actual characters, we can obtain successful and plausible searching behaviors in complex environments. The decomposition of the search task enabled us to reuse the single trained search policy for characters with different shapes and motor capabilities. Our key insight is that by depriving the privilege 3D information from the character, humanlike behaviors emerge because the character is forced to rely on its own egocentric vision perception and locomotion to complete the task. Furthermore, allowing the head of the character to move independently from the rest of the body leads to more natural searching behaviors, while facilitating the learning of a more effective policy. 

One limitation of our current system is the dependence on the mask channel to recognize and identify the target object. This assumption can be lifted by incorporating the state-of-the-art object recognition methods, such as \cite{he2017mask,xie2017aggregated}. In our current implementation, we chose to not incorporate memory structure in our policy beyond a very short history of the vision observations. We find that reasonable search behavior can be obtained without using memory for the set of environments we are working with. On the other hand, when working with more complex scenes, such as navigating in an entire building, memory becomes essential for localizing the agent and recognizing places that have been explored in the past. On the locomotion motion side, we notice that the collision checking during search policy learning sometimes is not sufficient when the policy is applied to the full character (Figure \ref{fig:failures}). A finer resolution collision checking might be needed to further improve the motion quality. Lastly, our scheme performs reasonably well for characters with relatively small variations in height during locomotion such as the ones presented here, however, the characters with more complex dynamics in the head motion or tasks that require active control of the character height might be of a challenge and further research on the topic is necessary.

There are a few promising future avenues for our work. First, is enabled interactions between the character and the environment, such as opening the fridge or drawers. This will allow the emergence of more intricate searching behaviors. Furthermore, our algorithm takes the character from a random location in the room to be in front of the object of interest. This provides an ideal initial pose for the character to perform downstream manipulation tasks such as pouring water into a cup, or pick up a phone. How to incorporate manipulation into our system and achieve more complex human behaviors is thus an important future work direction.

\appendix
\section{Hyperparameters}

In the Table. \ref{table:hyperparameters} we provide a complete list of the hyperparameters used for training the CURL.

\begin{table}[h]
\centering
\begin{tabular}{ | l | c | }
  \hline
  \textbf{Parameter} & \textbf{Setting} \\ \hline
  Image Size & 100x100 \\ \hline
  Augmentation & Random Crop (84x84) \\ \hline
  Image History Buffer Size & 5 \\ \hline
  Head History Buffer Size & 1 \\ \hline
  Replay buffer & 100000 \\ \hline
  Discount rate $\gamma$ & 0.99 \\ \hline
  Number of training steps & 0.75M \\ \hline
  Batch-size & 32 \\ \hline
  Alpha (SAC) : initial temperature & 0.1 \\ \hline
  Alpha (SAC) : learning rate & 0.0001 \\ \hline
  Alpha (SAC) : optimizer $\beta_1$ &  0.5 \\ \hline
  Alpha (SAC) : optimizer $\beta_2$ &  0.999 \\ \hline
  Actor : learning rate & 0.001 \\ \hline
  Actor : optimizer $\beta_1$ &  0.9 \\ \hline
  Actor : optimizer $\beta_2$ &  0.999 \\ \hline
  Actor : Number of layers & 4 \\ \hline
  Actor : Hidden dim & 1024 \\ \hline
  Actor : Activation function & ReLU \\ \hline
  Critic : learning rate & 0.001 \\ \hline
  Critic : $\tau$ (polyak averaging) & 0.01 \\ \hline
  Critic : optimizer $\beta_1$ &  0.9 \\ \hline
  Critic : optimizer $\beta_2$ &  0.999 \\ \hline
  Critic : Number of layers & 4 \\ \hline
  Critic : Hidden dim & 1024 \\ \hline
  Critic : Activation function & ReLU \\ \hline
  Encoder (CNN) : learning rate & 0.001 \\ \hline
  Encoder (CNN) : $\tau$ (polyak averaging) & 0.05 \\ \hline
  Encoder (CNN) : Number of layers & 4 \\ \hline
  Encoder (CNN) : Number of filters & 32 \\ \hline
  Encoder (CNN) : Latent dimension & 128 \\ \hline
  Encoder (CNN) : Activation function & ReLU \\ \hline
  
\end{tabular}
\caption{A complete set of CURL hyperparameters used to conduct all of the training experiments.}
\label{table:hyperparameters}



\end{table}

\bibliographystyle{eg-alpha-doi} 
\bibliography{paper}       


\end{document}